\documentclass[11pt]{article}

\usepackage[T1]{fontenc}
\usepackage[utf8]{inputenc}
\usepackage{lmodern}
\usepackage[a4paper,margin=25mm,headheight=15pt]{geometry}
\usepackage{microtype}
\usepackage{setspace}
\usepackage{amsmath,amssymb,mathtools}
\usepackage{graphicx}
\usepackage{booktabs,tabularx,array,longtable,multirow}
\usepackage{siunitx}
\usepackage{enumitem}
\usepackage{caption,subcaption}
\usepackage{float}
\usepackage[section]{placeins}
\usepackage{needspace}
\usepackage{etoolbox}
\usepackage{xcolor}
\usepackage{tikz}
\usetikzlibrary{arrows.meta,positioning,fit,calc,backgrounds,shapes.geometric,shapes.multipart}
\usepackage{titlesec}
\usepackage{fancyhdr}
\usepackage{hyperref}
\usepackage{orcidlink}
\usepackage[backend=biber,style=numeric-comp,sorting=none,maxbibnames=12]{biblatex}
\definecolor{Ink}{HTML}{1F2933}
\definecolor{Blue}{HTML}{245B78}
\definecolor{Sky}{HTML}{EAF1F5}
\definecolor{Teal}{HTML}{2B6F6D}
\definecolor{Coral}{HTML}{8A4B43}
\definecolor{Gold}{HTML}{8B6F2E}
\definecolor{Slate}{HTML}{52606D}
\definecolor{Mist}{HTML}{F7F8F9}
\definecolor{Rule}{HTML}{C8D0D7}

\hypersetup{
  colorlinks=true,
  linkcolor=Blue,
  citecolor=Blue,
  urlcolor=Teal,
  pdftitle={Zero-shot reasoning for simulating scholarly peer review},
  pdfauthor={Khalid M. Saqr},
  pdfsubject={A stand-alone operational and human-reference benchmark of the xPeer engine},
  pdfkeywords={peer review simulation, scholarly publishing, benchmarking, reproducibility, human-reference evaluation, zero-shot reasoning, xPeer, xPeerd, TRACE-R}
}
\setlist[itemize]{leftmargin=1.35em,itemsep=0.25em,topsep=0.35em}
\titleformat{\section}{\Large\bfseries\color{Ink}}{\thesection}{0.65em}{}
\titleformat{\subsection}{\large\bfseries\color{Ink}}{\thesubsection}{0.6em}{}
\titleformat{\subsubsection}{\normalsize\bfseries\color{Ink}}{\thesubsubsection}{0.55em}{}
\titlespacing*{\section}{0pt}{2.0em}{0.6em}
\titlespacing*{\subsection}{0pt}{1.5em}{0.45em}
\pretocmd{\section}{\Needspace{7\baselineskip}}{}{}
\pretocmd{\subsection}{\Needspace{5\baselineskip}}{}{}
\newcolumntype{Y}{>{\raggedright\arraybackslash}X}
\newcommand{\xpeer}{xPeer}
\newcommand{\xpeerd}{xPeerd}

\begin{document}
\pagenumbering{arabic}
\title{\bfseries Zero-shot reasoning for simulating scholarly peer review}
\author{Khalid M. Saqr\textsuperscript{\orcidlink{0000-0002-3058-2705}}\\
\normalsize College of Engineering and Technology\\
\normalsize Arab Academy for Science, Technology, and Maritime Transport\\
\normalsize Alexandria 1029-- EGYPT}
\date{}

\maketitle

\begin{abstract}
\noindent Scholarly publishing requires scalable scrutiny supported by auditable evidence. This paper presents a two-component benchmark of \xpeer{}, the peer-review simulation engine delivered through the \xpeerd{} web front-end. The operational component analyzes 352 of 500 simulation records retained under stable-task criteria across disciplines and review modes. The human-reference component releases 1,108 version-1 F1000Research manuscript records with linked human reports and applies a prespecified two-human/two-\xpeer{} comparison. Human-review text and recommendations remained outside the generation input, and source joining occurred after \xpeer{} outputs had been persisted. This procedure defines workflow-level review withholding; prior model exposure falls outside the recorded design. Among 802 records with exactly two human reports, 271 contained two usable \xpeer{} reviewer fields, giving a complete-pair availability rate of 33.8\%. Under deterministic extraction rules, median manuscript-level report length was 1,889 words for \xpeer{} and 763 for humans, while median concern count was 41 and 13, respectively. \xpeer{} reports showed higher targeting, category coverage, and executability. Human reports showed higher explicit-reasoning language, lexical manuscript attestation, taxonomy-based scientific relevance, and lower mean within-source redundancy. Cross-source lexical concern matching and recommendation agreement were low. The evidence therefore defines distinct observable review profiles and a transparent reproducibility baseline. Scientific correctness of individual concerns, autonomous editorial use, and cross-system superiority require expert adjudication and common-protocol testing. The study-level dataset and exact version-pinned reproducibility record are archived on Zenodo \cite{Knowdyn2026Benchmark,Knowdyn2026BenchmarkVersion}.
\end{abstract}


\section{Introduction}
Peer review remains a capacity-constrained form of scholarly quality control. A major survey documented researchers' views of the benefits, burdens, and possible alternatives to conventional peer review \cite{Ware2008ElsevierStudy}. An author-perspective study separately examined how review duration relates to perceived review quality \cite{Huisman2017AcceptanceRates}. Proposals to pay reviewers and regulate publication volume make the capacity problem explicit, although they remain policy arguments rather than evaluations of a particular intervention \cite{LSeghier2025}.

Editorial practice is also changing. eLife first tested a revised consultation process and later adopted a publishing model centered on reviewed preprints and public assessments \cite{eLife2016Trial,eLifeNewModel2023}. Nature Portfolio and JAMA Network Open document distinct venue-specific editorial and peer-review workflows \cite{NaturePublishingProcess,JAMAOpenPeerReview}. These developments motivate evaluating review-simulation systems both in operational workflows and against manuscript-linked human reports.

AI-assisted reviewing covers several different tasks that should not be conflated. MetaWriter studies AI support for writing within scientific peer review \cite{Sun2024}. Other systems predict quality or acceptance and recommend reviewers \cite{Pendyala2025}, or combine human and model-derived information to assess methodological novelty \cite{Wu2025}. One study evaluates whether ChatGPT can predict review outcomes across platforms \cite{Thelwall2025}. Separate evaluations test automatic paper reviewing and document limits in reliability and specificity \cite{Zhou20249340,Carabantes2023}. A survey synthesizes the automated scholarly-review literature \cite{Zhuang2025}, and a medical informatics article reviews generative AI in manuscript peer review \cite{Hoyt2025}. An ethics analysis focuses specifically on disclosure, confidentiality, accountability, and publisher procedures for AI-assisted reviewing \cite{Mollaki2024239}.

The design space is similarly heterogeneous. AgentReview uses LLM agents to simulate peer-review dynamics \cite{Jin20241208}. Peer-review aspects have been used to guide scientific-article summarization rather than to produce a full editorial review \cite{Majadly2025537}. Other publications discuss rapid editorial screening and discipline-specific use of ChatGPT-4 in peer review \cite{Fiorillo2024,Biswas2024441}. Editorial commentary describes AI-assisted peer review as support under human accountability \cite{NatureBiomedEng2024665,Crawford2024}. Journal-focused commentary separately reviews the challenges and opportunities of AI in editorial practice \cite{Shah2024196}. Critiques of generative AI in publishing emphasize authorship, disclosure, and institutional-integrity risks \cite{Rahimi2024110,Mehta2024}. Assessment-focused commentary questions whether AI can validly judge research quality \cite{Ryan2024S18}.

Changes in manuscript production form a related but distinct literature. Surgical and medical commentaries discuss whether and how models should contribute to scientific writing and authorship \cite{Keller2023751,Gutierrez2023281}. A cardiology review examines the use of language models in medical-science writing \cite{Bhattaru20241950}. Empirical work has tested whether reviewers can distinguish human-written from ChatGPT-written abstracts \cite{Levin2024669}. Specialty-specific articles consider generative AI in obstetrics, gynecology, and journal practice \cite{Kawakita2025,Hillard2025301}, while a separate experiment asks whether AI-generated scientific discussion can pass journal peer review \cite{Sheridan2025}.

The evaluation landscape now includes multi-model behavioral studies, multidimensional benchmarks, and large operational deployments. PRAIB compares about 11,000 generated reviews from five models with human reports on 1,000 ICLR and NeurIPS papers \cite{Zurawicki2026PRAIB}. PRISM evaluates five automated reviewer systems and human reviewers across depth, novelty assessment, flaw identification and prioritization, and constructiveness \cite{Loc2026PRISM}. The AAAI-26 pilot generated an identified AI review for each of 22,977 full-review submissions within a conference process that retained human editorial authority \cite{Biswas2026AAAI}. One survey organizes automated scholarly-review capabilities \cite{Zhuang2025}; a later survey distinguishes generation, post-review tasks, and human-centered, reference-based, model-based, and aspect-oriented evaluation \cite{Wu2026Survey}.

The reasoning literature also requires narrower attribution. Chain-of-thought prompting showed that worked intermediate steps can improve performance when demonstrations are supplied \cite{Wei2022CoT}. Zero-shot chain-of-thought instead elicited intermediate reasoning through an instruction without task-specific exemplars \cite{Kojima2022}. Plan-and-Solve prompting added explicit problem decomposition to that zero-shot setting \cite{Wang20232609}, whereas the CoT Collection improved zero- and few-shot behavior through reasoning-oriented fine-tuning \cite{Kim202312685}.

Structured zero-shot methods have then been studied in several task-specific settings. MedAgents uses role-based LLM collaboration for zero-shot medical reasoning \cite{Tang2024599}, while LLMs have also been evaluated as zero-shot models of people in human--robot interaction \cite{Zhang20237961}. Visual question answering has been approached through reasoning prompts and frozen language models \cite{Lan20234389,Guo202310867}. Separate studies address zero-shot object navigation \cite{Dorbala20244083}, composed-image retrieval \cite{Yang202480}, and reward modeling for reinforcement learning \cite{Rocamonde2024}. These results motivate structured zero-shot review simulation, but none directly establishes scientific correctness for full-manuscript review.

The surrounding research-tool ecosystem increasingly combines language models with retrieval, specialized representations, and agent orchestration. Scientific hypothesis generation has been tested directly \cite{Park2024578}; literature-based discovery and evidence-based-science applications have been reviewed separately \cite{Taleb2024,Zhang2025}. Research-synthesis and assistant systems include domain-specific fine-tuning \cite{Susnjak2025}, general research-assistant prototypes \cite{Gheorghe2024}, a modular bioengineering chatbot \cite{Forootani2025}, and deep web research with reasoning models \cite{Li2025WebThinker}.

Domain-grounded implementations likewise address different problems. Retrieval-augmented systems have been developed for construction-safety guidance and multi-agent pharmacovigilance \cite{Baek2025,Choi2024}. Biological discovery services combine transformer models with retrieval \cite{Kim2024}, while other systems target battery-science knowledge defragmentation \cite{Zhao2025} and multimodal materials-science workflows \cite{Katzer2025}. Explainable and agentic methods have been used for biomedical knowledge synthesis \cite{Pelletier2025}. Separate work addresses LLM-assisted electrocatalyst discovery \cite{Shen20251921} and knowledge synthesis for biomanufacturing \cite{Li2025}. In the humanities, LLM--RAG has been applied to knowledge-graph construction from oral-history archives \cite{Sun2024142}.

Methodological work identifies more specific design choices. Fact-centric retrieval structures information around claims \cite{Sinha2024421}, while SiReRAG studies how indexing similar and related information affects multihop reasoning \cite{Zhang202599376}. Multi-agent GraphRAG has been evaluated in an e-government assistant \cite{Papageorgiou2025}. Domain-specific embeddings have been developed for accelerator physics \cite{Hellert2025}, and scientific knowledge-graph work has studied semantic similarity and neuro-symbolic discovery \cite{Nguyen2025,Schmidt2024}. These studies concern tasks other than peer review, but they motivate treating grounding, representation, and orchestration as separately testable components.

At a conceptual level, computational models of scientific reasoning provide one foundation for automated review \cite{Thagard1988}. Work on cognition and design cautions that computational artifacts participate in human practices rather than merely replacing them \cite{Winograd1986}, while models of epistemic communities analyze how communication structure affects collective inquiry \cite{Zollman2007}. Knowledge representation supplies formal tools for encoding claims and relations \cite{Sowa2000}. Argument-mining surveys describe methods for identifying argumentative components and structures in text \cite{Lippi2016,Stede2018}. Parsing work provides a concrete implementation for recovering argument structure from persuasive essays \cite{Gurevych2017}. Together, these traditions support a distinction central to this benchmark: textual structure can be measured computationally, whereas the validity and importance of a scientific criticism require accountable expert assessment.

\xpeer{} was developed as a zero-shot reasoning engine for structured scholarly scrutiny. \xpeerd{} provides the web workflow through which authors, publishers, and conferences access the engine \cite{xPeerdWeb2026}. The benchmark reported here treats the system as an operational research object and evaluates its outputs through two complementary components. The operational component measures disciplinary coverage, task-conditioned behavior, simulated decisions, issue load, and procedural anchoring. The human-reference component draws on the Re3 family of resources, which aligns scientific revisions with associated reviews and extends that material in Re3-Sci2.0 \cite{Ruan2024Re3,Ruan2024Re3Sci2}. This benchmark then applies its own version-1 and linked-report eligibility rules, with human-review content held outside the generation workflow until \xpeer{} outputs have been persisted.

The study addresses four research questions. First, how does \xpeer{} behave across disciplines and review tasks? Second, how do paired human and \xpeer{} reports differ in scale, concern structure, targeting, reasoning language, manuscript attestation, category coverage, executability, relevance, and redundancy? Third, how strongly do the two sources correspond at concern and recommendation levels? Fourth, how completely can the resource construction, cohort decisions, statistical analysis, and outputs be reproduced?

The contribution is a self-contained benchmark architecture that joins operational evidence with a public same-manuscript comparison. Its claims are expressed at the level supported by the released data: report structure, detector-recognized concern patterns, source correspondence, analytical availability, and reproducibility. Expert adjudication remains the required basis for scientific correctness and comparative system ranking.

\section{Study overview and benchmark positioning}

\begin{figure}[!htbp]
\centering
\begin{tikzpicture}[
  node distance=7mm,
  card/.style={rounded corners=2mm,draw=Rule,line width=0.8pt,fill=white,
               text width=0.225\textwidth,minimum height=31mm,align=left,inner sep=3.2mm,font=\small},
  num/.style={circle,fill=Ink,text=white,font=\bfseries\footnotesize,minimum size=6.5mm,inner sep=0pt},
  arr/.style={-{Latex[length=2.5mm]},line width=1.0pt,draw=Blue}
]
\node[card] (s1) {\textbf{Operational\\component}\\[1mm]
352 valid simulations\\
Cross-field and task\\variation\\
Decisions, workload, and anchoring};
\node[card,right=of s1] (s2) {\textbf{Human-reference\\component}\\[1mm]
1,108 public benchmark\\records\\
271 manuscripts in strict cohort\\
Two human and two \xpeer{} reports};
\node[card,right=of s2] (s3) {\textbf{Integrated\\interpretation}\\[1mm]
Operational behavior\\
Same-manuscript\\comparison\\
Multidimensional,\\auditable evidence};
\node[num] at ([xshift=7mm,yshift=3mm]s1.north west) {1};
\node[num] at ([xshift=7mm,yshift=3mm]s2.north west) {2};
\node[num] at ([xshift=7mm,yshift=3mm]s3.north west) {3};
\draw[arr] (s1.east) -- (s2.west);
\draw[arr] (s2.east) -- (s3.west);
\begin{scope}[on background layer]
\node[fit=(s1)(s2)(s3),rounded corners=3mm,fill=Mist,draw=Rule,inner sep=5mm] {};
\end{scope}
\end{tikzpicture}
\caption{Benchmark architecture. The operational component quantifies system behavior across fields and tasks. The human-reference component supplies a public same-manuscript comparison. The integrated interpretation separates operational breadth, analytical availability, source correspondence, and scientific-validity boundaries. The 1,108 records denote persisted HTTP-success records; 271 exact-two-human cases contain two usable \xpeer{} reviewer fields.}
\label{fig:evidence-architecture}
\end{figure}

The benchmark evaluates a system through documented evidence layers: operational inputs, persisted outputs, cohort construction, textual observables, statistical uncertainty, and reproducibility controls. Table~\ref{tab:benchmark-at-glance} summarizes the two components and their analytical roles.

\begin{table}[!htbp]
\centering
\caption{Benchmark components and analytical roles.}
\label{tab:benchmark-at-glance}
\small
\begin{tabularx}{\textwidth}{@{}p{0.18\textwidth}YY@{}}
\toprule
 & \textbf{Operational component} & \textbf{Human-reference component}\\
\midrule
Primary purpose & Characterize operational breadth and task-conditioned behavior. & Compare human and \xpeer{} scrutiny on the same manuscript version.\\
Starting resource & 500 operational simulation records. & 1,146 eligible version-1 manuscripts with at least two linked human reports.\\
Analytic set & 352 valid stable-task reports. & 1,108 released records; 271 strict paired manuscripts.\\
Comparison object & Variation across subject groups and review modes. & Two human reports and two \xpeer{} reports per manuscript.\\
Main evidence & Decisions, issue load, classification, task behavior, and page anchoring. & Report scale, concern units, TRACE-R profile, category prevalence, overlap, redundancy, and recommendations.\\
Reproducibility & Public analysis code; source reports supplied under provider permission. & Public versioned dataset, notebooks, exclusions, hashes, quality criteria, and machine-readable outputs.\\
Interpretive role & Operational capability and workflow behavior. & Principal source-comparison evidence.\\
\bottomrule
\end{tabularx}
\end{table}

\subsection{Position in the 2026 evaluation landscape}
Peer-review AI studies now span behavioral benchmarking, direct multi-system evaluation, and live deployment. Table~\ref{tab:landscape} compares documented protocol properties. Performance ranking requires shared manuscripts, common prompts, expert-adjudicated outcomes, and aligned cost and latency measures.

\begin{table}[!htbp]
\centering
\caption{Representative peer-review AI evidence designs available by 2026.}
\label{tab:landscape}
\scriptsize
\begin{tabularx}{\textwidth}{@{}>{\raggedright\arraybackslash}p{0.15\textwidth}>{\raggedright\arraybackslash}p{0.20\textwidth}>{\raggedright\arraybackslash}p{0.21\textwidth}YY@{}}
\toprule
\textbf{Study or system} & \textbf{Scale and context} & \textbf{Human-reference design} & \textbf{Distinctive contribution} & \textbf{Comparison scope}\\
\midrule
PRAIB \cite{Zurawicki2026PRAIB} & 1,000 ICLR/NeurIPS papers; about 11,000 generated reviews from five models & Generated and human reviews compared across behavior, style, and engagement observables & Large multi-model behavioral benchmark & Conference-domain corpus and study-specific metrics\\
PRISM \cite{Loc2026PRISM} & Five automated reviewer systems and human reviewers on a stratified ICLR/ICML/NeurIPS corpus & Multidimensional depth, novelty, flaw-identification, prioritization, and constructiveness assessment & Direct cross-system multidimensional comparison & Study-specific corpus, scoring framework, and verification design\\
AAAI-26 pilot \cite{Biswas2026AAAI} & 22,977 full-review conference submissions in a live deployment & AI reviews supplied as an identified additional input; humans retained decisions & Conference-scale operational feasibility and user evaluation & Deployment-scale and workflow evidence\\
\xpeer{} benchmark & 1,108 multidisciplinary journal records; 271 complete two-human/two-\xpeer{} cases & Same version-1 manuscript; two human and two \xpeer{} reports; workflow-level review withholding & Public resource, concern-level traceability, explicit attrition, and reproducible tests & Single-system evaluation with expert adjudication reserved for future comparative testing\\
\bottomrule
\end{tabularx}
\end{table}

\section{Evaluated system and scope}
\subsection{System identity and workflow}
\xpeer{} is the peer-review simulation engine. \xpeerd{} is the web front-end available at \href{https://xpeerd.com}{xPeerd.com}. The benchmark evaluates engine outputs, while the front-end provides the submission and delivery workflow. Retained notebook-generated figures use the legacy label \emph{xPeerd}; within those figures, the label refers to outputs generated by the \xpeer{} engine through the \xpeerd{} workflow.

The system produces structured scrutiny for pre-submission, editorial, and reviewer-support settings. Its target output includes two simulated reviewer reports, an editorial summary, and a recommendation field. The review modes cover conventional critique, data-analysis review, conference review, repeated-review simulation, and double-blind simulation. The benchmark evaluates field coverage, task-conditioned behavior, issue structure, manuscript targeting, concern-category representation, revision executability, source correspondence, and recommendation association.

\begin{figure}[p]
\centering
\includegraphics[height=0.8\textheight,keepaspectratio]{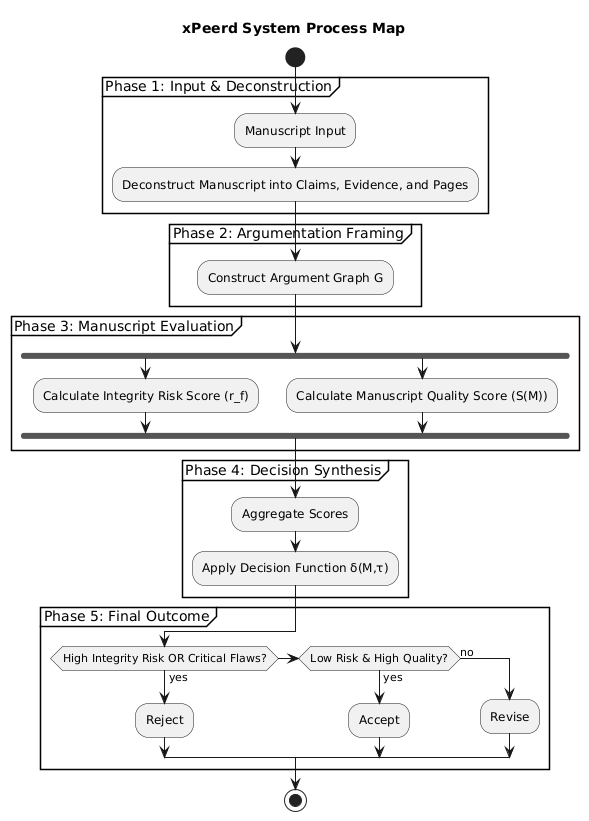}
\caption{System process map used in the operational benchmark design. The retained figure summarizes manuscript deconstruction, argument framing, evaluation, decision synthesis, and final output. The formal Methods section identifies the design-level constructs and the variables directly observed in the released benchmark.}
\label{fig:s1-process-map}
\end{figure}

\subsection{Evaluation scope and evidence claims}
The benchmark evaluates \xpeer{} through a multidisciplinary resource, manuscript-linked human reports, workflow-level review withholding, explicit analytical attrition, concern-level traceability, uncertainty analysis, and machine-readable quality controls. The empirical claims concern observable outputs and reproducible analyses. Scientific correctness, calibrated editorial judgment, latency, cost, and cross-system rank require dedicated expert-adjudicated experiments.

The analysis retains each TRACE-R dimension separately and reports both favorable and unfavorable source differences. This profile-based design preserves the distinction between review breadth, explicit rationale, manuscript attestation, executability, relevance, redundancy, and recommendation correspondence.

\section{Materials and methods}
\subsection{Formal design and evaluated implementation of xPeer}
The benchmark formalizes zero-shot peer-review simulation at three levels: system design, evaluated implementation, and empirical observables. This separation links each equation to a defined role in the study.

At design level, a manuscript is represented as
\begin{equation}
M=\langle C,E,P\rangle,
\label{eq:manuscript-design}
\end{equation}
where $C$ is a set of manuscript claims, $E$ is a set of evidential units such as text passages, tables, and figures, and $P$ is the available location index. The task selector is
\begin{equation}
\tau\in\{\mathrm{HCReview},\mathrm{DAReview},\mathrm{confReview},\mathrm{PRR},\mathrm{DBReviewSim}\}.
\label{eq:task-design}
\end{equation}
For construction of the human-reference resource, the submitted task was \texttt{/DBReviewSim}. The intended output is a structured review object
\begin{equation}
\delta(M,\tau)=\langle R_1,R_2,S_{\mathrm{ed}},L\rangle,
\label{eq:structured-output}
\end{equation}
where $R_1$ and $R_2$ are simulated reviewer reports, $S_{\mathrm{ed}}$ is an editorial summary, and $L$ is a decision or recommendation field. A guard set constrains execution:
\begin{equation}
\mathcal{G}(M,\tau)=\{\mathrm{scope},\mathrm{manuscript\ present},\mathrm{evidence\ targeting},\mathrm{task\ format}\}.
\label{eq:guard-set}
\end{equation}
A failed guard triggers a correction request or refusal. Fabrication of missing input falls outside the admissible operational behavior.

Bayesian updating, argumentation graphs, integrity-risk functions, and weighted manuscript scores remain design-level abstractions described in the framework literature \cite{Saqr2025xPeer}. The released analysis uses the submitted task, persisted response fields, parsed review text, and downstream observables. Priors, likelihood functions, Dung extensions, fraud probabilities, and calibrated decision thresholds were outside the measured variable set.

Human-reference submissions were executed through the \xpeer{} service during July 2026 using up to ten concurrent workers, a maximum of eight HTTP attempts for retryable responses, and persisted per-case output before human reports were joined. The construction notebook specified a long request timeout and a fixed \texttt{/DBReviewSim} command. The public dataset omits the underlying model version and generation-temperature controls, leaving those factors unstratified. Authentication credentials were supplied through an environment variable and remain outside the article package and benchmark record.

\subsection{Study design and notation}
The following notation defines the paired design and its six separately reported TRACE-R dimensions.

Let $i\in\{1,\ldots,N\}$ index manuscripts, let $s\in\{H,X\}$ denote source, with $H$ for human and $X$ for \xpeer{}, and let $r\in\{1,2\}$ index the two reports from each source. The report set for manuscript $i$ is
\begin{equation}
\mathcal{R}_i=\{R_{iHr}:r=1,2\}\cup\{R_{iXr}:r=1,2\}.
\label{eq:report-set}
\end{equation}
Here, $R_{isr}$ is the text of reviewer $r$ from source $s$ for manuscript $i$.

A record entered the strict cohort when it contained exactly two nonempty human reports and exactly two usable \xpeer{} reviewer reports. With $n_{is}$ denoting the usable report count for source $s$ and $m_i$ indicating the presence of nonempty manuscript text, the inclusion indicator was
\begin{equation}
I_i=\mathbf{1}(n_{iH}=2)\,\mathbf{1}(n_{iX}=2)\,\mathbf{1}(m_i=1),
\label{eq:inclusion}
\end{equation}
where $\mathbf{1}(\cdot)$ equals one when its condition is true and zero otherwise. The strict cohort size was $N=\sum_i I_i=271$.

\subsection{Operational-component observables}
For the operational component, each valid report $j$ was assigned an All Science Journal Classification supergroup and a classification confidence $p_j\in[0,1]$. A report passed the declared confidence rule when
\begin{equation}
C^{\mathrm{class}}_j=\mathbf{1}(p_j\geq\tau),\qquad \tau=0.20.
\label{eq:s1-classification}
\end{equation}
The page-anchor fraction for report $j$ was the number of extracted issues containing a page reference, $a_j$, divided by the total issue count, $m_j$:
\begin{equation}
q_j=\frac{a_j}{\max(1,m_j)}.
\label{eq:s1-anchor-fraction}
\end{equation}
Anchoring compliance was defined as
\begin{equation}
C^{\mathrm{anchor}}_j=\mathbf{1}(q_j\geq0.20).
\label{eq:s1-anchor-rule}
\end{equation}
These quantities describe operational classification and anchoring behavior. Scientific-accuracy inference requires external adjudication.

\subsection{Human-review-withheld benchmark construction}
The source corpus was Re3-Sci2.0, which includes scientific document revisions and associated reviews from F1000RD and NLPeer \cite{Ruan2024Re3Sci2}. For the present benchmark, F1000Research records were restricted to version 1 and required at least two linked human reports. Manuscript inputs contained the title, abstract, manuscript text, source metadata, and identifiers. Human review text, reference reviewer metadata, recommendation labels, and decisions remained in a separate store. \xpeer{} transport outputs were written to persistent storage before joining on document identifiers. This procedure defines the workflow-level withholding control; prior model exposure to public source material remained unmeasured. The released resource contains 1,108 persisted HTTP-success records and is archived through the study-level and version-pinned Zenodo records \cite{Knowdyn2026Benchmark,Knowdyn2026BenchmarkVersion}.

\subsection{Concern-unit extraction}
A concern unit was a sentence of at least five words that matched at least one declared condition: explicit concern language, an explicit revision action, a question form, or concern/recommendation section context without praise-only language. Let $U_{isr}=\{u_{isrk}\}_{k=1}^{K_{isr}}$ be the extracted unit set for report $R_{isr}$, where $K_{isr}$ is the number of units.

Let $W(R_{isr})$ be the number of word tokens in report $R_{isr}$. Report-level concern density per 1,000 words was
\begin{equation}
D_{isr}=1000\frac{K_{isr}}{\max\{1,W(R_{isr})\}}.
\label{eq:concern-density}
\end{equation}

Each unit was assigned to the first matching category in the ordered taxonomy: statistics, study design, methods and reproducibility, data and results, interpretation and claims, literature context, ethics and reporting, presentation and clarity, or other scientific.

\subsection{Measurement validity, parser audit, and source-style effects}
The released benchmark schema contains four \xpeer{} fields: Reviewer~1, Reviewer~2, editorial summary, and recommendation. Analytical usability required both reviewer fields to be nonempty. Among the 802 exact-two-human records, 531 contained fewer than two usable reviewer fields. Table~\ref{tab:parser-states} reports their packaged structural states. Raw API payloads and transport metadata are absent from the release, leaving incomplete generation, heading variation, and parser allocation as unresolved mechanisms. The paired cohort uses complete fields as stored, with zero imputation.

The concern detector uses explicit lexical and structural rules. Such rules can interact with source style: a templated system report may expose headings, action verbs, and target cues more readily, whereas a human report may express the same content implicitly. A blinded manual annotation subset was unavailable for source-specific precision, recall, category accuracy, and inter-annotator agreement. Consequently, $K$, $T$, $G$, $C$, $E$, and $V$ are treated as detector-dependent observables. Latent-quality validation remains a future expert-annotation task. Sixty-one reports yielded zero concern units: 22 human reports and 39 \xpeer{} reports. They were retained with zero-valued unit-derived features.

\subsection{TRACE-R observables}
For unit $u_{isrk}$, let $t_{isrk}$ equal one when an explicit manuscript target or location is named; let $g_{isrk}$ equal one when explicit rationale language is present; let $e_{isrk}$ equal one when an actionable revision verb is present; and let $v_{isrk}$ equal one when the unit belongs to a category beyond presentation and clarity.

Report-level targeting, explicit reasoning, executability, and relevance were the corresponding unit proportions:
\begin{align}
T_{isr}&=\frac{1}{\max(1,K_{isr})}\sum_{k=1}^{K_{isr}}t_{isrk},
\label{eq:targeting}\\
G_{isr}&=\frac{1}{\max(1,K_{isr})}\sum_{k=1}^{K_{isr}}g_{isrk},
\label{eq:reasoning}\\
E_{isr}&=\frac{1}{\max(1,K_{isr})}\sum_{k=1}^{K_{isr}}e_{isrk},
\label{eq:executability}\\
V_{isr}&=\frac{1}{\max(1,K_{isr})}\sum_{k=1}^{K_{isr}}v_{isrk}.
\label{eq:relevance}
\end{align}

To estimate attested alignment, each manuscript was divided into overlapping word chunks. For unit $u_{isrk}$, let $z_{isrk}$ be the maximum TF--IDF cosine similarity to any manuscript chunk and let $o_{isrk}$ be the proportion of content tokens in the unit that occur in its best-matching chunk. The unit alignment score was
\begin{equation}
a_{isrk}=0.75z_{isrk}+0.25o_{isrk}.
\label{eq:unit-alignment}
\end{equation}
The report-level attested alignment was
\begin{equation}
A_{isr}=\frac{1}{\max(1,K_{isr})}\sum_{k=1}^{K_{isr}}a_{isrk}.
\label{eq:attested-alignment}
\end{equation}

Let $B_{isr}$ be the number of distinct categories present in report $R_{isr}$ and let $J=9$ be the number of taxonomy categories. Coverage was
\begin{equation}
C_{isr}=\frac{B_{isr}}{J}.
\label{eq:coverage}
\end{equation}

For each manuscript and source, the two reports were aggregated. Word count and concern count were summed,
\begin{equation}
W_{is}=\sum_{r=1}^{2}W(R_{isr}),\qquad K_{is}=\sum_{r=1}^{2}K_{isr},
\label{eq:source-count-aggregation}
\end{equation}
while density and TRACE-R proportions were averaged:
\begin{equation}
\bar{Z}_{is}=\frac{1}{2}\sum_{r=1}^{2}Z_{isr},\qquad Z\in\{D,T,G,A,C,E,V\}.
\label{eq:source-aggregation}
\end{equation}
In the main text, explicit reasoning is denoted $R$, and relevance is denoted $R_2$ for mnemonic presentation; Equations~\ref{eq:reasoning} and \ref{eq:relevance} use $G$ and $V$ to avoid symbol collision.

\subsection{Within-source redundancy and cross-source matching}
For the two reports from source $s$, concern units were compared by TF--IDF cosine similarity using unigram and bigram features. One-to-one assignments maximized total similarity. With $d_{is}$ accepted unit pairs at threshold $\eta$ and report unit counts $K_{is1}$ and $K_{is2}$, within-source redundancy was
\begin{equation}
Q_{is}=\frac{d_{is}}{\max\{1,\min(K_{is1},K_{is2})\}}.
\label{eq:redundancy}
\end{equation}
The default threshold was $\eta=0.35$; sensitivity analyses used 0.25, 0.30, 0.35, 0.40, 0.45, and 0.50.

Human and \xpeer{} units for manuscript $i$ were also assigned one-to-one. If $m_i(\eta)$ pairs exceeded threshold $\eta$, the source-normalized matched fractions were
\begin{equation}
M_{iH}(\eta)=\frac{m_i(\eta)}{\max(1,K_{iH})},\qquad
M_{iX}(\eta)=\frac{m_i(\eta)}{\max(1,K_{iX})},
\label{eq:matched-fractions}
\end{equation}
where $K_{iH}$ and $K_{iX}$ are total concern counts across the two reports from each source. These fractions quantify lexical correspondence. Validity and novelty require expert adjudication.

\subsection{Category prevalence}
For category $c$, manuscript $i$, and source $s$, the binary presence indicator was
\begin{equation}
Y_{isc}=\mathbf{1}\!\left(\sum_{r=1}^{2}\sum_{k=1}^{K_{isr}}\mathbf{1}[u_{isrk}\in c]>0\right).
\label{eq:category-presence}
\end{equation}
Source prevalence was the mean of $Y_{isc}$ across the 271 manuscripts. Paired source differences were tested with exact McNemar tests and Benjamini--Hochberg false-discovery-rate correction. Jaccard overlap and phi correlation were reported separately.

\subsection{Recommendation normalization}
Human recommendations were mapped from source metadata. \xpeer{} recommendations were mapped from explicit recommendation, decision, or verdict language and assigned high, medium, or none confidence. The ordinal encoding was reject $=0$, revise or reservations $=1$, and approve $=2$. For manuscripts with usable source consensus, the mean source recommendation was
\begin{equation}
L_{is}=\frac{1}{n^{\mathrm{rec}}_{is}}\sum_{r\in\mathcal{O}_{is}}\ell_{isr},
\label{eq:recommendation-consensus}
\end{equation}
where $\ell_{isr}\in\{0,1,2\}$, $\mathcal{O}_{is}$ is the set of reports with observable labels, and $n^{\mathrm{rec}}_{is}=|\mathcal{O}_{is}|$. Rounded exact agreement was
\begin{equation}
P_{\mathrm{exact}}=\frac{1}{N_{\mathrm{rec}}}\sum_{i=1}^{N_{\mathrm{rec}}}\mathbf{1}\!\left(\operatorname{round}L_{iH}=\operatorname{round}L_{iX}\right).
\label{eq:recommendation-agreement}
\end{equation}
Recommendation analyses also reported Spearman association, Lin concordance, quadratic weighted kappa, and mean absolute ordinal error.

\subsection{Paired inference and agreement}
For any manuscript-level observable $Z$, the paired difference was
\begin{equation}
\Delta_i(Z)=Z_{iX}-Z_{iH}.
\label{eq:paired-difference}
\end{equation}
The analysis reported source means and medians, mean paired differences, bootstrap 95\% confidence intervals, paired rank-biserial effect sizes, paired Wilcoxon tests, and sign-flip permutation tests. Benjamini--Hochberg correction was applied within declared test families.

Association and agreement were intentionally separated. Spearman correlation assessed rank association. Lin's concordance correlation coefficient assessed absolute agreement. Distance correlation, mean absolute error, root-mean-square error, and Bland--Altman bias and limits were included as complementary diagnostics. These tests quantify association, agreement, and error. Individual-concern correctness remains an expert-adjudication outcome.

\subsection{Reproducibility and quality control}
The benchmark-construction notebook verified the upstream archive SHA-256, expected source counts, input schema, uniqueness of benchmark identifiers, and separation of human-review data from submitted payloads. The TRACE-R notebook verified the released package SHA-256, cohort counts, balanced report rows, concern-unit coverage, metric bounds, test completeness, category reconciliation, recommendation auditing, output existence, and figure export. The machine-readable completion report recorded 22 passed checks out of 22. These checks establish computational consistency and artifact completeness. Scientific validity depends on construct validation and domain adjudication.

The version-pinned input archive was \texttt{xpeerd\_benchmark\_study\_2026\_v1.0.0.zip} with SHA-256 \texttt{0ab7cb88d2b2db687b586ad303e017a9db0a8104e10f1b8e18c30b8f6a75129c}. The random seed was 20260723. Confidence intervals used 2,000 bootstrap replicates; permutation procedures used 1,999 replicates. Manuscript chunks contained 160 words with a 40-word overlap, the minimum concern-unit length was five words, and the primary cross-source matching threshold was 0.35 with sensitivity analysis from 0.25 to 0.50.

The article-package validation environment used Python 3.13.5, pandas 2.2.3, NumPy 2.3.5, SciPy 1.17.0, scikit-learn 1.8.0, Matplotlib 3.10.8, and Seaborn 0.13.2. The public reproducibility package defines its supported environment through versioned environment files, output manifests, and hashes. Repository state was pinned at commit \texttt{99e602873ddb1f7dca8a08d8aa05979e1fce643e}, dated 23 July 2026 \cite{xPeerEvaluationRepo2026}.

\section{Operational benchmark results}
The operational component used 500 simulation records generated between February and May 2023. Stable-task inclusion retained 352 reports, corresponding to 70.4\% of the starting set. Records outside the analytic set represented off-task, misfired, or free-form interactions. The component characterizes system behavior across disciplines and review modes; scientific accuracy is evaluated only through subsequent expert-adjudicated work.

\subsection{Disciplinary and task breadth}
The retained reports spanned Life Sciences, Physical Sciences, Health Sciences, Social Sciences, and Humanities. Physical Sciences and Health Sciences formed the largest groups, with 109 and 113 reports respectively; Humanities contributed 70, Social Sciences 46, and Life Sciences 14. No report was assigned to the multidisciplinary fallback class. Normalized assignment confidence averaged 0.45 with a standard deviation of approximately 0.12, and every retained report exceeded the declared 0.20 threshold. The operational data covered conventional critique, data-analysis review, double-blind simulation, and repeated-review simulation; the analytic set contained no valid conference-review cases.

\begin{figure}[!htbp]
\centering
\includegraphics[width=1\textwidth]{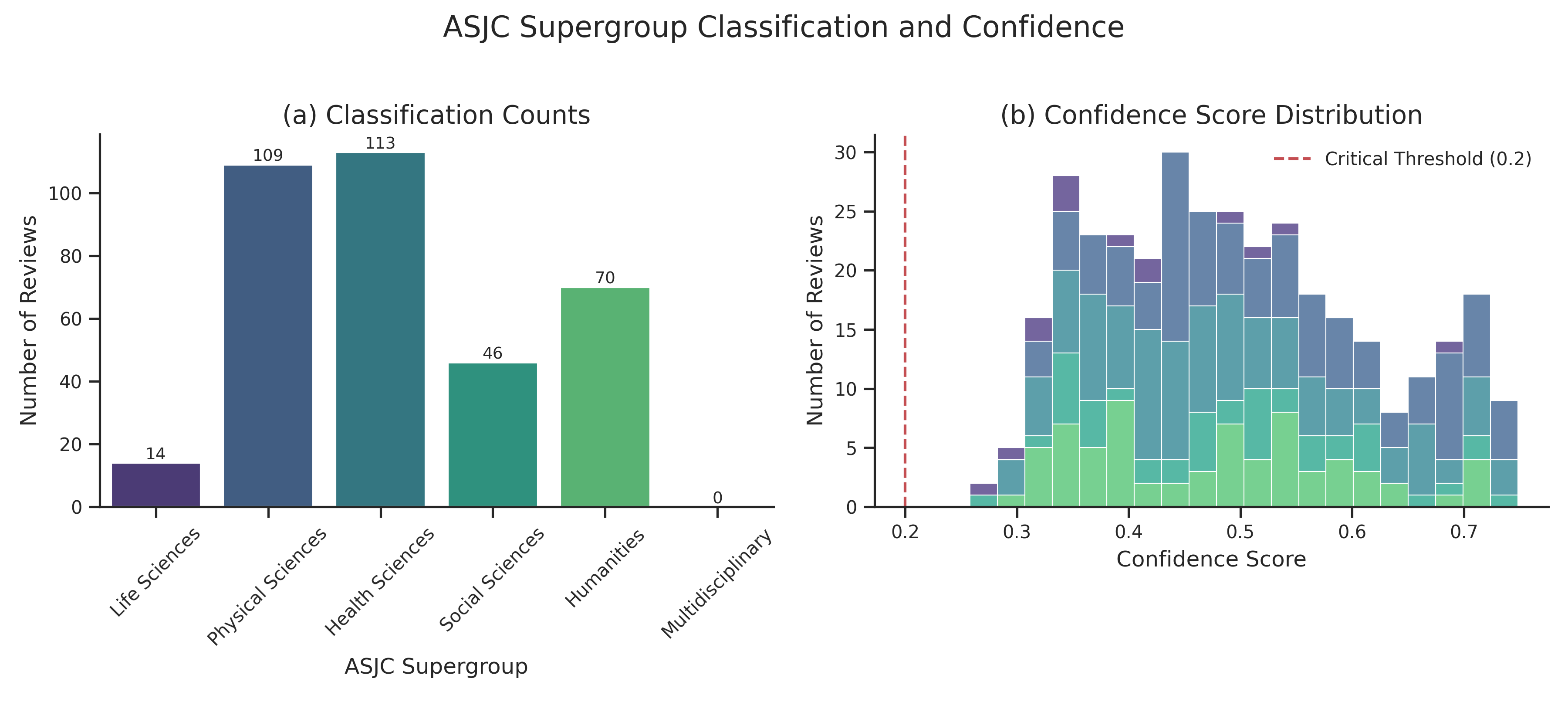}
\caption{Subject classification and confidence for 352 valid operational reports. The figure quantifies disciplinary coverage within the stable-task analytic set.}
\label{fig:s1-classification}
\end{figure}

\subsection{Simulated decisions and issue load}
Revision formed more than half of simulated outcomes in every represented disciplinary group. Rejection was approximately 42\% in Life Sciences and 45\% in Health Sciences, while Physical Sciences and Humanities remained below 20\%. Acceptance was approximately 3--12\% in Health Sciences, Social Sciences, and Humanities and was negligible in Life Sciences and Physical Sciences. These distributions quantify field-conditioned decision behavior within the operational records. Calibration to venue decisions requires manuscript-matched editorial outcome data.

\begin{figure}[!htbp]
\centering
\begin{subfigure}[t]{0.49\textwidth}
\includegraphics[width=\linewidth]{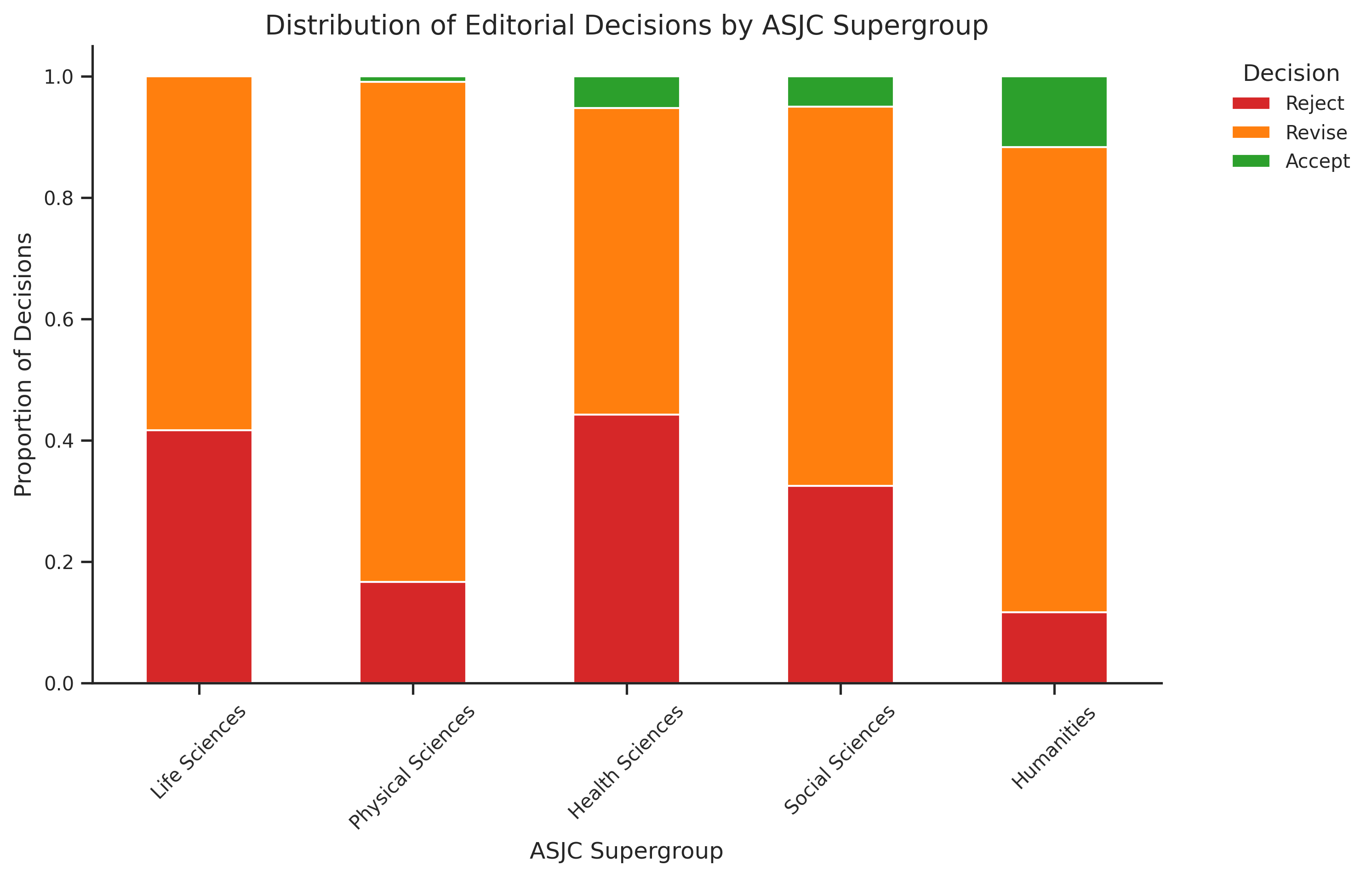}
\caption{Editorial decision composition by subject group.}
\end{subfigure}\hfill
\begin{subfigure}[t]{0.49\textwidth}
\includegraphics[width=\linewidth]{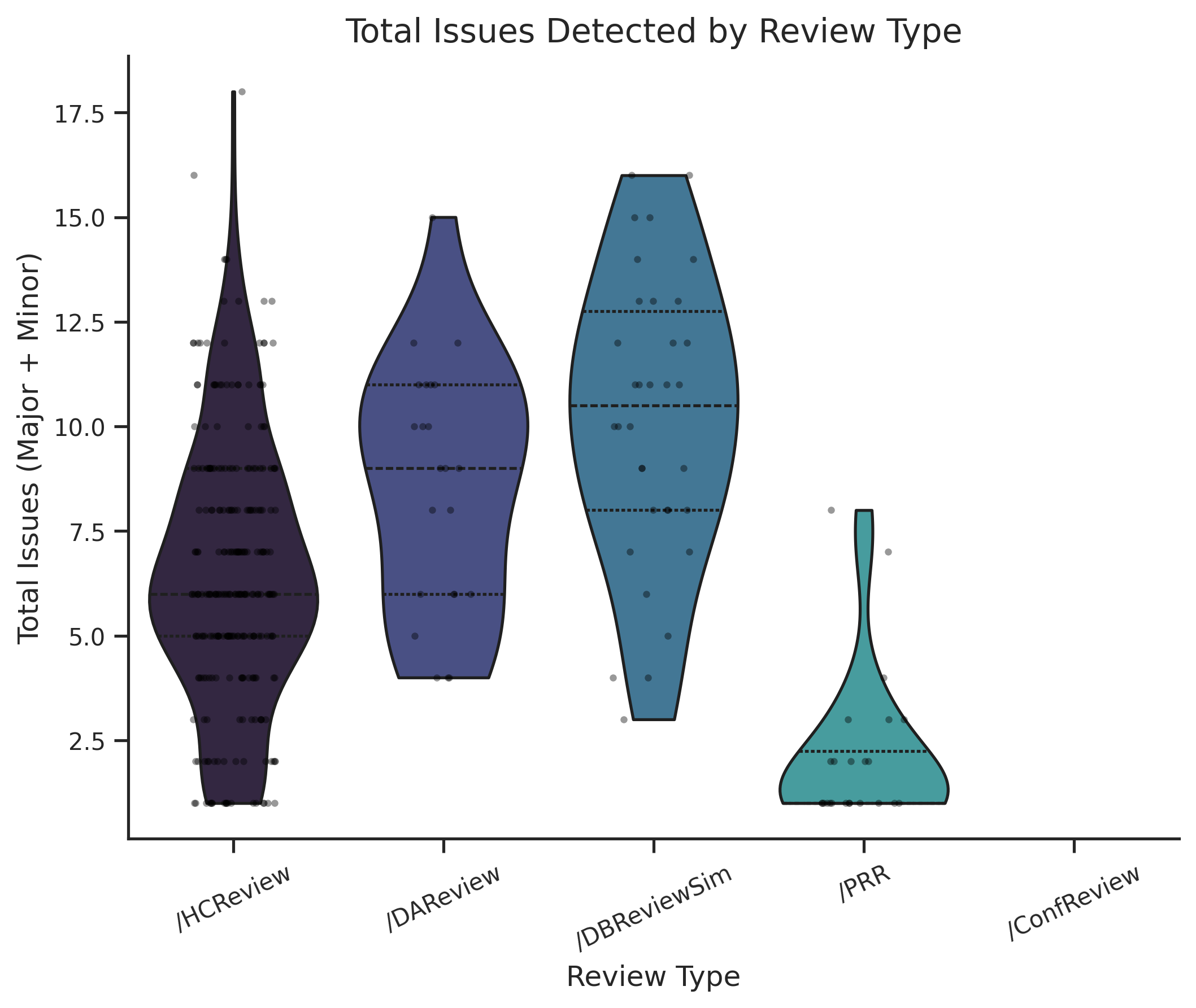}
\caption{Issue-count distributions by review mode.}
\end{subfigure}
\caption{Operational decision and workload results. The two panels show variation in simulated outcomes and critique volume across subjects and review tasks.}
\label{fig:s1-decision-workload}
\end{figure}

Issue-load distributions varied by review type. Conventional critique reports were concentrated at roughly 4--10 issues, with an upper range near 18. Data-analysis reports were centered higher, with approximate quartiles of 6--12 issues. Double-blind simulation produced the widest spread, from about 3 to more than 16 issues, with a median near 10. Repeated-review simulation was concentrated at 1--3 terminal issues and rarely exceeded 8. The result quantifies sensitivity to task design. Issue count is an output-volume measure whose scientific utility depends on the validity and priority of individual concerns.

\subsection{Procedural anchoring}
The benchmark measured the fraction of detected issues containing page references and its relationship with report length. Report length had a weak positive association with page-anchor fraction (Spearman $\rho=0.13$, $p=0.014$). The overall mean compliance rate under the declared threshold was 0.29. Approximate disciplinary means were 0.34 for Physical Sciences, 0.30 for Social Sciences, 0.29 for Life Sciences and Health Sciences, and 0.20 for Humanities. Data-analysis and repeated-review modes were near 0.35, while conventional critique and double-blind simulation were near 0.28--0.30. These results quantify partial manuscript-location anchoring and its variation across fields and tasks.

\begin{figure}[!htbp]
\centering
\begin{subfigure}[t]{0.49\textwidth}
\includegraphics[width=\linewidth]{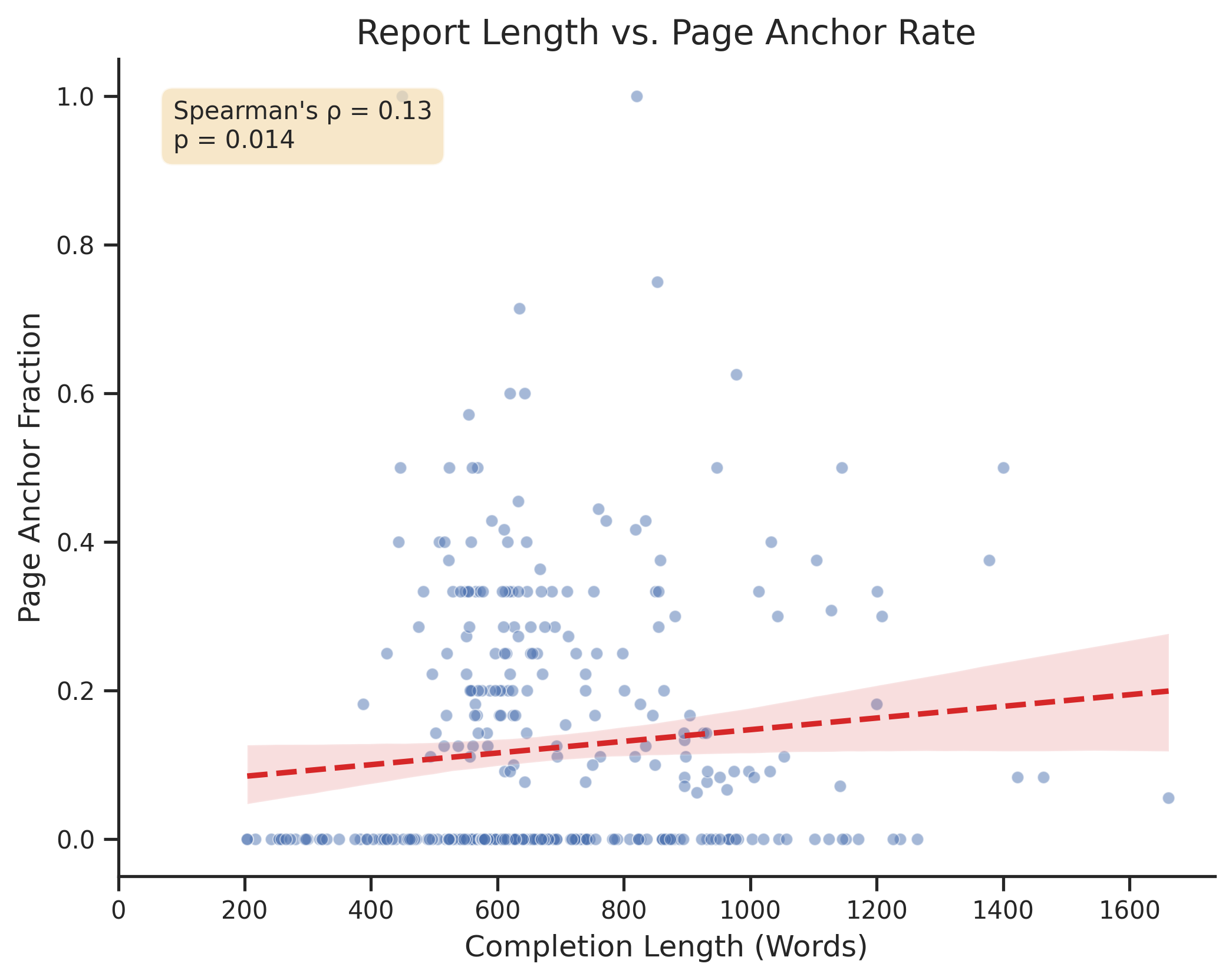}
\caption{Report length and page-anchor fraction.}
\end{subfigure}\hfill
\begin{subfigure}[t]{0.49\textwidth}
\includegraphics[width=\linewidth]{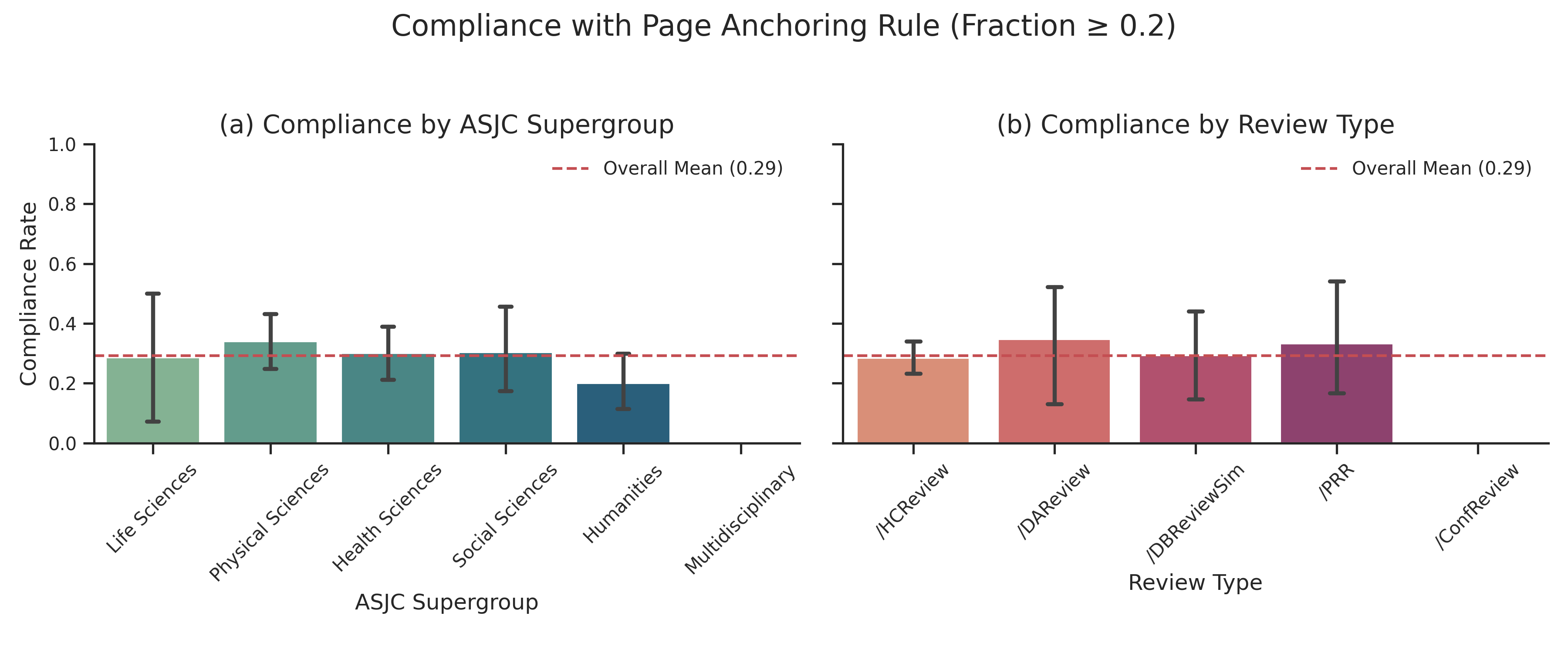}
\caption{Page-anchoring compliance by field and review type.}
\end{subfigure}
\caption{Operational anchoring results. Page references were observable across fields and tasks, with a mean issue-level anchor fraction of 0.29.}
\label{fig:s1-anchoring}
\end{figure}

\clearpage
\begin{table}[!htbp]
\centering
\caption{Systematic summary of the operational benchmark.}
\label{tab:operational-summary}
\small
\begin{tabularx}{0.94\textwidth}{@{}p{0.29\textwidth}Y@{}}
\toprule
\textbf{Outcome} & \textbf{Result}\\
\midrule
Stable-task inclusion & 352 of 500 reports (70.4\%); 148 off-task, misfired, or free-form interactions were excluded.\\
Disciplinary classification & Life Sciences 14; Physical Sciences 109; Health Sciences 113; Social Sciences 46; Humanities 70; mean normalized confidence 0.45 (SD approximately 0.12), with all retained reports above 0.20.\\
Decision behavior & Revision exceeded 50\% in every field; rejection was approximately 42\% in Life Sciences and 45\% in Health Sciences; acceptance was rare.\\
Task-conditioned workload & Conventional critique concentrated at 4--10 issues; data-analysis reports at approximately 6--12; double-blind simulation had the widest spread and a median near 10; repeated-review simulation concentrated at 1--3 terminal issues.\\
Procedural anchoring & Mean issue-level page-anchor compliance 0.29; report length association $\rho=0.13$ ($p=0.014$); field means approximately 0.20--0.34 and task means approximately 0.28--0.35.\\
Inference scope & Operational breadth, task response, simulated workload, and traceability behavior.\\
\bottomrule
\end{tabularx}
\end{table}

\paragraph{Interpretive scope.}
The operational component supports claims about system coverage, task differentiation, simulated decision patterns, issue volume, and anchoring behavior. Same-manuscript correspondence and source-profile differences are estimated in the human-reference component.

\section{Human-reference resource and analytical cohort}
\subsection{Resource yield}
Re3-Sci2.0 supplies the underlying scientific revisions and associated reviews \cite{Ruan2024Re3Sci2}. Application of the present benchmark's prespecified eligibility criteria identified 1,146 F1000Research manuscripts and 2,661 version-linked human reports. Each eligible record represented manuscript version 1 and had at least two linked human reports; these counts and eligibility conditions are outputs of the benchmark-construction procedure, not results reported by the source study.

Of the 1,146 eligible submissions, 1,108 produced persisted HTTP-success records and entered release 1.0.0 \cite{Knowdyn2026Benchmark,Knowdyn2026BenchmarkVersion}. Human-review text and recommendation data remained outside the submitted inputs and were joined only after \xpeer{} responses had been persisted. This control establishes workflow-level review withholding, but it cannot exclude prior model exposure to public source material. Analytical usability was assessed separately because persisted records could contain incomplete reviewer fields. The package SHA-256 is \texttt{0ab7cb88d2b2db687b586ad303e017a9db0a8104e10f1b8e18c30b8f6a75129c}.

\begin{figure}[!htbp]
\centering
\begin{tikzpicture}[
  node distance=9mm and 6mm,
  topbox/.style={rounded corners=1.8mm,draw=Rule,line width=0.8pt,fill=white,
                 text width=0.17\textwidth,minimum height=19mm,align=center,inner sep=2.4mm,font=\small},
  lowbox/.style={rounded corners=1.8mm,draw=Rule,line width=0.8pt,fill=white,
                 text width=0.17\textwidth,minimum height=17mm,align=center,inner sep=2.4mm,font=\small},
  arr/.style={-{Latex[length=2.3mm]},line width=0.95pt,draw=Blue},
  hold/.style={-{Latex[length=2.3mm]},line width=0.95pt,dashed,draw=Coral}
]
\node[topbox] (source) {\textbf{Re3-Sci2.0}\\1,780 source documents};
\node[topbox,right=of source] (eligible) {\textbf{Eligible version 1}\\1,146 manuscripts\\2,661 linked human reports};
\node[topbox,right=of eligible] (safe) {\textbf{Review-withheld input}\\manuscript content\\and metadata};
\node[topbox,right=of safe] (persist) {\textbf{Persisted \xpeer{} output}\\1,108 HTTP-success records};
\draw[arr] (source) -- (eligible);
\draw[arr] (eligible) -- (safe);
\draw[arr] (safe) -- (persist);

\node[lowbox,below=13mm of eligible,draw=Coral,fill=Gold!5] (human) {\textbf{Held-out human reports}\\separate reference store};
\node[lowbox,below=13mm of safe,fill=Sky!35] (join) {\textbf{Post-persistence join}\\records linked by identifier};
\node[lowbox,below=13mm of persist,fill=Sky!60] (strict) {\textbf{Strict paired cohort}\\271 manuscripts\\2 human + 2 \xpeer{} reports};
\draw[hold] (eligible.south) -- (human.north);
\draw[hold] (human.east) -- (join.west);
\draw[arr] (persist.south) -- (strict.north);
\draw[arr] (join.east) -- (strict.west);
\begin{scope}[on background layer]
\node[fit=(source)(eligible)(safe)(persist)(human)(join)(strict),rounded corners=3mm,fill=Mist,draw=Rule,inner sep=5mm] {};
\end{scope}
\end{tikzpicture}
\caption{Resource provenance, workflow-level review withholding, and strict paired cohort. Human reports remained in a separate reference store until \xpeer{} outputs had been persisted. The released resource contains 1,108 HTTP-success records, of which 271 exact-two-human cases contain two usable reviewer fields.}
\label{fig:benchmark-flow}
\end{figure}

\subsection{Cohort accounting and analytical availability}
The 1,108-record resource supports future parsing and evaluation. The strict cohort supports a balanced comparison in which each manuscript contributes exactly two reports from each source.

Exclusions were mutually exclusive. The resource contained 265 records with three human reviews, 33 with four, and 8 with five. Among 802 records with exactly two human reviews, 421 contained zero usable \xpeer{} reviewer fields and 110 contained one. The remaining 271 manuscripts formed the strict cohort, corresponding to a complete-pair availability rate of $271/802=33.8\%$. This level of attrition creates a material complete-case selection risk and is carried through the interpretation.

Packaged-field inspection characterized the 531 incomplete exact-two-human records. Among the 421 zero-reviewer cases, 125 had all four parsed fields empty, 294 stored nonempty text only in the recommendation field, and 2 stored text only in the editorial-summary and recommendation fields. Among the 110 one-reviewer cases, 106 contained Reviewer~1 only and 4 contained Reviewer~2 only. These states are compatible with blank output, heading variation, section-allocation failure, and incomplete two-reviewer structure. Raw transport payloads were outside the public release, leaving the underlying mechanisms unresolved. The paired analysis used the released fields exactly as stored, with zero imputation, reconstruction, or reassignment.

\begin{table}[!htbp]
\centering
\caption{Human-reference cohort accounting.}
\label{tab:cohort}
\small
\begin{tabularx}{0.88\textwidth}{@{}Y r@{}}
\toprule
\textbf{Cohort state or mutually exclusive exclusion} & \textbf{Records}\\
\midrule
Released benchmark records & 1,108\\
Exclude: three human reports & 265\\
Exclude: four human reports & 33\\
Exclude: five human reports & 8\\
Exactly two human reports before \xpeer{} usability filter & 802\\
Exclude: zero usable \xpeer{} reviewer reports & 421\\
Exclude: one usable \xpeer{} reviewer report & 110\\
\textbf{Include: exactly two human and two usable \xpeer{} reports} & \textbf{271}\\
\bottomrule
\end{tabularx}
\end{table}

\begin{table}[!htbp]
\centering
\caption{Structural states among 531 exact-two-human records with incomplete \xpeer{} reviewer fields.}
\label{tab:parser-states}
\small
\begin{tabularx}{0.92\textwidth}{@{}Y r@{}}
\toprule
\textbf{Observed packaged-field state} & \textbf{Records}\\
\midrule
All four parsed fields empty & 125\\
Text only in Recommendation & 294\\
Text in Editorial summary and Recommendation & 2\\
Reviewer 1 present; Reviewer 2 absent & 106\\
Reviewer 2 present; Reviewer 1 absent & 4\\
\textbf{Total incomplete exact-two-human records} & \textbf{531}\\
\bottomrule
\end{tabularx}
\end{table}

\subsection{TRACE-R analytical coverage}
TRACE-R comprises Targeting, explicit Reasoning, Attested alignment, category Coverage, Executability, and scientific Relevance. The dimensions are reported separately as observable text properties. Additional diagnostics cover report length, concern density, within-source redundancy, cross-source lexical matching, category prevalence, and recommendation correspondence.

Concern extraction produced 15,563 units across 1,023 of 1,084 reports, giving report-unit coverage of 94.4\%. Sixty-one reports contained zero extracted units: 22 of 542 human reports and 39 of 542 \xpeer{} reports. They remained in the analysis with zero-valued unit-derived observables. The report balance was exact at 542 human reports and 542 \xpeer{} reports. The detector uses lexical and structural rules, and the two sources differ in style and formatting; inference is therefore confined to detector-dependent observables.

\section{Comparative benchmark results}
Primary outcomes comprised report scale, concern count and density, and the six TRACE-R observables. Category prevalence, cross-source matching, redundancy, and recommendation correspondence served as secondary diagnostics. Statistical uncertainty quantifies paired source differences conditional on the declared measurement rules. Construct validity is limited by the absence of a blinded, source-stratified expert-annotation study.

\subsection{Report scale and concern volume}
At manuscript level, the median combined length of the two \xpeer{} reports was 1,889 words, compared with 763 words for the two human reports. The mean paired difference was 940 words, with a bootstrap 95\% confidence interval from 833 to 1,040 words. The paired rank-biserial effect size was 0.836.

The median extracted concern count was 41 for \xpeer{} and 13 for humans. The mean paired difference was 25.4 concerns, with a bootstrap 95\% confidence interval from 22.9 to 27.9 and a paired rank-biserial effect size of 0.904. Concern density was also higher for \xpeer{}, indicating a difference beyond report length alone.

\begin{figure}[!htbp]
\centering
\includegraphics[width=\textwidth]{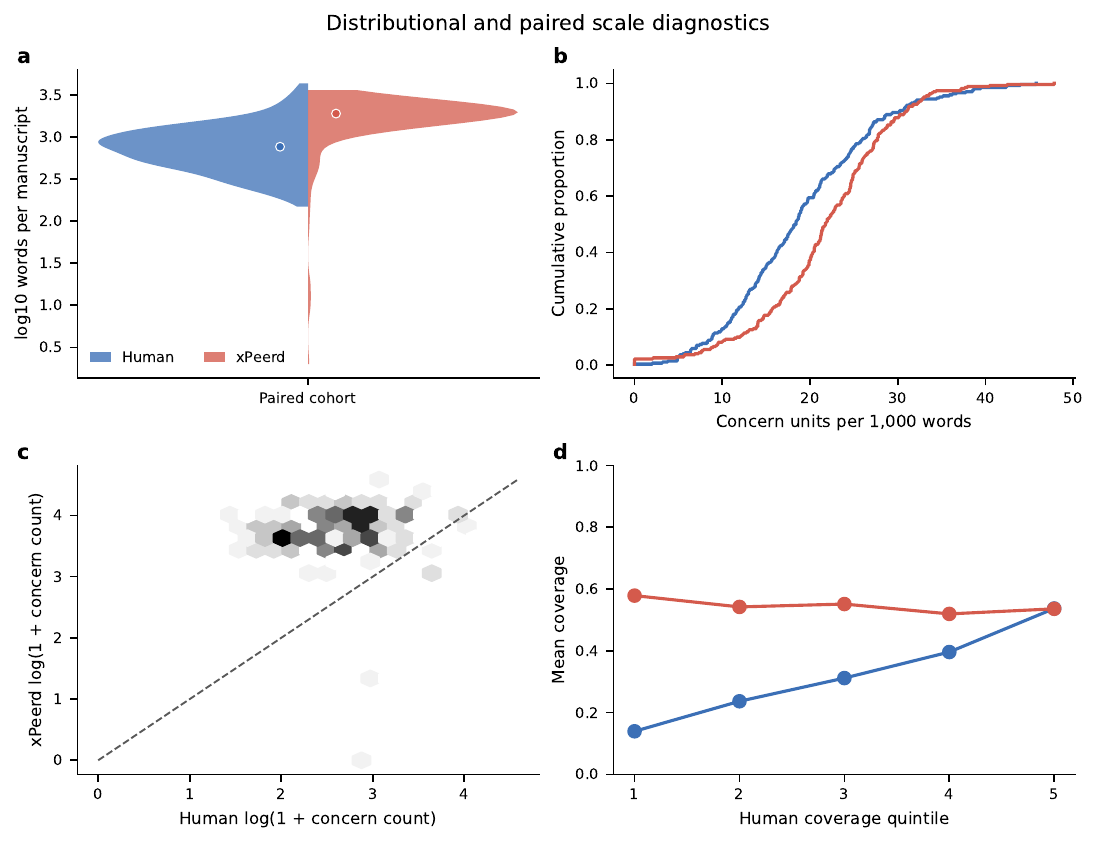}
\caption{Distributional and paired scale diagnostics for the strict cohort. The panels compare report length, concern density, concern counts, and category coverage.}
\label{fig:s2-scale}
\end{figure}

The paired distributions and effect sizes show a systematic scale difference across manuscripts. The detector recognizes concern, question, action, and section-context patterns. The result therefore quantifies detector-recognized scrutiny; scientific-defect accuracy requires expert adjudication.

\subsection{Multidimensional TRACE-R profile}
\xpeer{} showed higher values on three primary observables. Targeting was 0.414 for \xpeer{} and 0.279 for humans; category coverage was 0.546 and 0.324; executability was 0.669 and 0.606. These differences indicate more explicit manuscript targets, broader representation of the prespecified taxonomy, and more revision-action language in the \xpeer{} reports.

Human reports showed higher explicit-reasoning language, with 0.081 compared with 0.018; attested alignment, with 0.119 compared with 0.096; and scientific relevance, with 0.932 compared with 0.886. The explicit-reasoning difference was the largest negative paired TRACE-R effect for \xpeer{}. The profile identifies a design priority: high-impact concerns require a clear inferential bridge from observation to consequence and requested action.

Within-source redundancy was 0.063 for \xpeer{} and 0.037 for humans. Both values were low, while the paired direction indicates more repeated content between the two simulated reviewers.

\begin{figure}[!htbp]
\centering
\includegraphics[width=\textwidth]{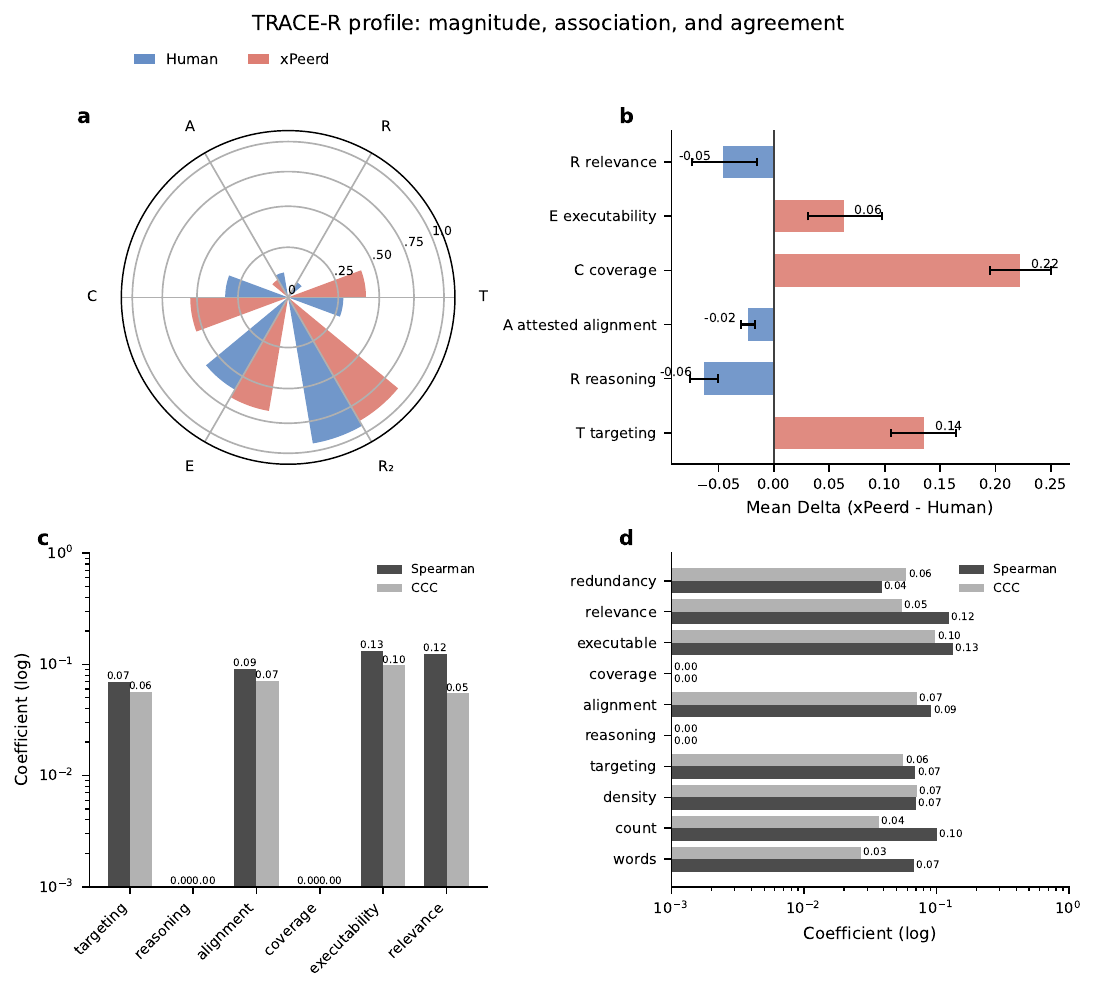}
\caption{TRACE-R profile, paired effects, cross-source association, and absolute agreement. Each dimension is reported separately.}
\label{fig:s2-trace}
\end{figure}

\begin{table}[!htbp]
\centering
\caption{Primary paired results for the strict 271-manuscript cohort. Values are source-level manuscript aggregates.}
\label{tab:primary-results}
\scriptsize
\begin{tabularx}{\textwidth}{@{}Y r r r r r@{}}
\toprule
\textbf{Observable} & \textbf{Human} & \textbf{\xpeer{}} & \textbf{Difference} & \textbf{95\% CI} & \textbf{Rank-biserial}\\
\midrule
Median report words & 763 & 1,889 & 940.5$^{a}$ & 832.9 to 1,040.3 & 0.836\\
Median concern count & 13 & 41 & 25.4$^{a}$ & 22.9 to 27.9 & 0.904\\
Mean concerns per 1,000 words & 18.981 & 21.597 & 2.616 & 1.395 to 3.921 & 0.262\\
Mean targeting & 0.279 & 0.414 & 0.135 & 0.105 to 0.164 & 0.557\\
Mean explicit reasoning & 0.081 & 0.018 & -0.063 & -0.076 to -0.051 & -0.720\\
Mean attested alignment & 0.119 & 0.096 & -0.023 & -0.030 to -0.017 & -0.525\\
Mean category coverage & 0.324 & 0.546 & 0.222 & 0.195 to 0.250 & 0.829\\
Mean executability & 0.606 & 0.669 & 0.063 & 0.031 to 0.098 & 0.270\\
Mean relevance & 0.932 & 0.886 & -0.046 & -0.074 to -0.015 & -0.319\\
Mean redundancy & 0.037 & 0.063 & 0.026 & 0.005 to 0.046 & 0.340\\
\bottomrule
\multicolumn{6}{@{}l}{\footnotesize $^{a}$Mean difference; the displayed source values for words and concern count are medians.}
\end{tabularx}
\end{table}

\subsection{Scientific-category prevalence}
The category analysis records whether at least one concern from a prespecified scientific category appears in the paired reports for a manuscript. \xpeer{} showed higher prevalence in statistics, study design, methods and reproducibility, data and results, interpretation and claims, literature context, ethics and reporting, and presentation and clarity. All reported differences remained significant after false-discovery-rate correction.

The largest gaps occurred in interpretation and claims, study design, methods and reproducibility, and presentation and clarity. Methods/reproducibility concerns appeared in 95.6\% of \xpeer{} manuscript pairs and 58.3\% of human pairs; data/results concerns appeared in 95.2\% and 72.0\%, respectively. The pattern quantifies detector-recognized breadth. Report length, templated structure, explicit headings, and action-oriented wording may contribute to the observed prevalence differences.

\begin{figure}[!htbp]
\centering
\includegraphics[width=\textwidth]{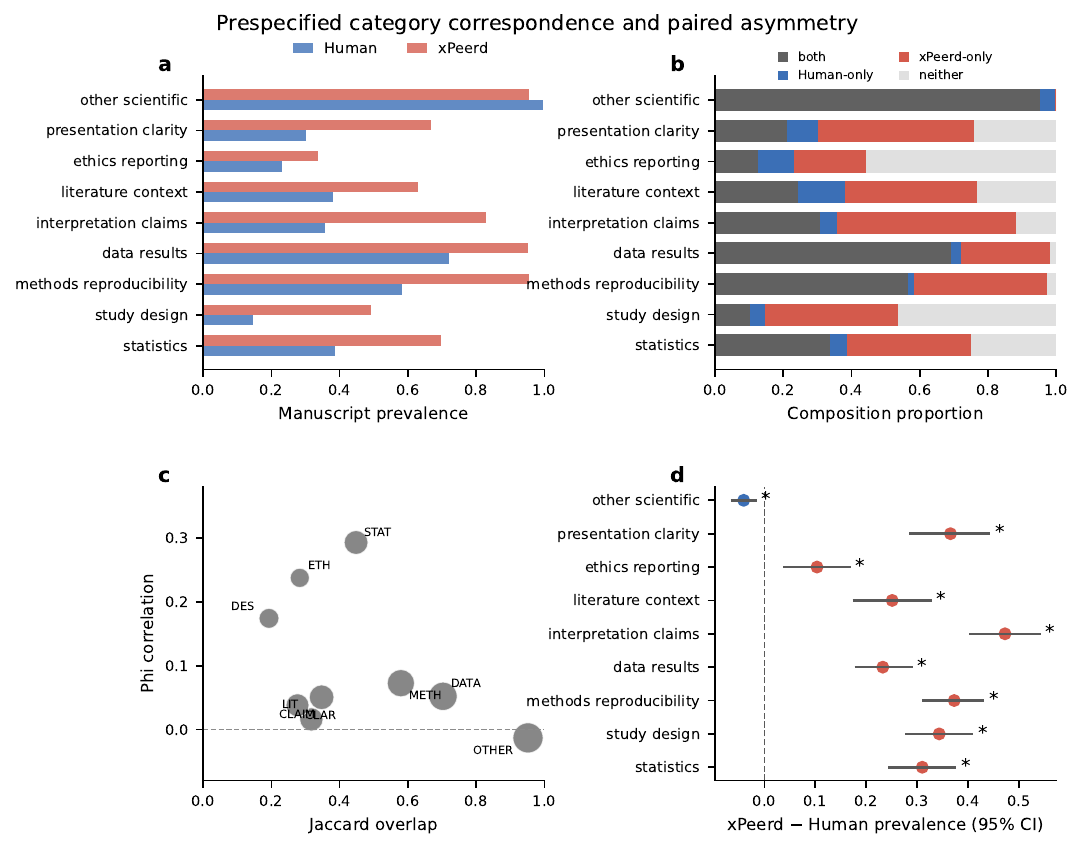}
\caption{Category prevalence, paired asymmetry, overlap, and exact paired tests. Higher prevalence represents broader detector-recognized coverage.}
\label{fig:s2-categories}
\end{figure}

\subsection{Cross-source concern correspondence}
Human and \xpeer{} concern units were assigned one-to-one and accepted above a prespecified lexical similarity threshold. The median matched fraction was zero in both source-normalized views. Mean human recovery was 0.026 and mean \xpeer{} alignment was 0.009. Fifty-six manuscripts produced at least one accepted pair for calculation of mean accepted similarity.

The low matched fractions define lexical non-equivalence between the source profiles. Plausible contributors include differentiated critical focus, alternative wording, concern decomposition, presentation-oriented comments, and weakly justified expansion. Expert adjudication is required to classify each unmatched unit by correctness, severity, relevance, and revision utility.

\begin{figure}[!htbp]
\centering
\includegraphics[width=\textwidth]{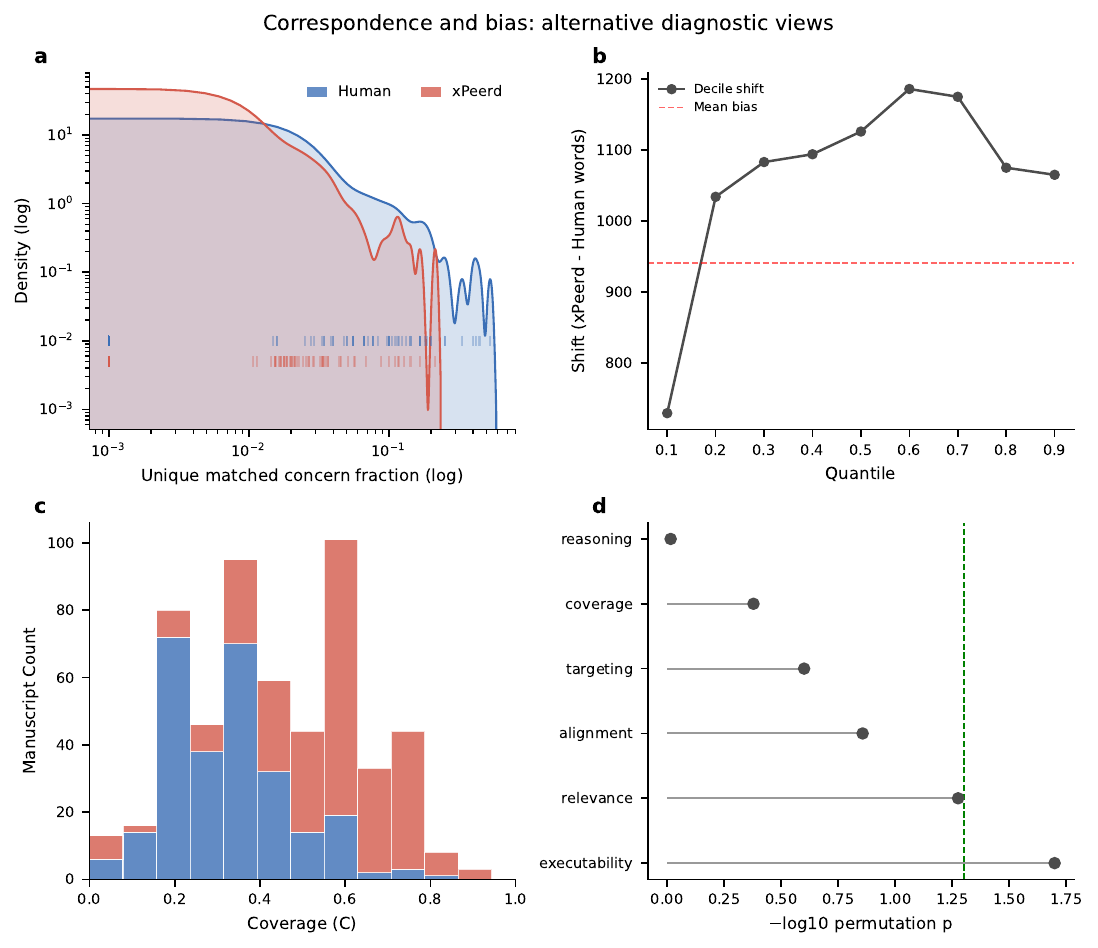}
\caption{Matched concern fractions, report-length shifts, coverage distributions, and permutation evidence.}
\label{fig:s2-diagnostics}
\end{figure}

Absolute agreement between source-level profiles was also low. Spearman association was small for most observables, and Lin concordance coefficients were close to zero. The supported inference is source non-equivalence. Value assessment requires scientific adjudication of the differentiated concerns.

\subsection{Recommendation correspondence}
Human recommendation metadata were available for all 542 human reports. Normalized recommendation language was extracted from 380 of 542 \xpeer{} reports, giving 70.1\% report-level coverage. At manuscript level, 240 cases had usable source consensus values. Rounded exact agreement was 43.75\%; Spearman association was 0.170; Lin concordance was 0.164; quadratic weighted kappa was 0.137; and mean ordinal error was 0.465.

The recommendation results indicate small positive association, low concordance, and incomplete system-label observability. Editorial decision use therefore remains under accountable human authority. The principal benchmark evidence concerns report structure and critique observables.

\begin{figure}[!htbp]
\centering
\includegraphics[width=\textwidth]{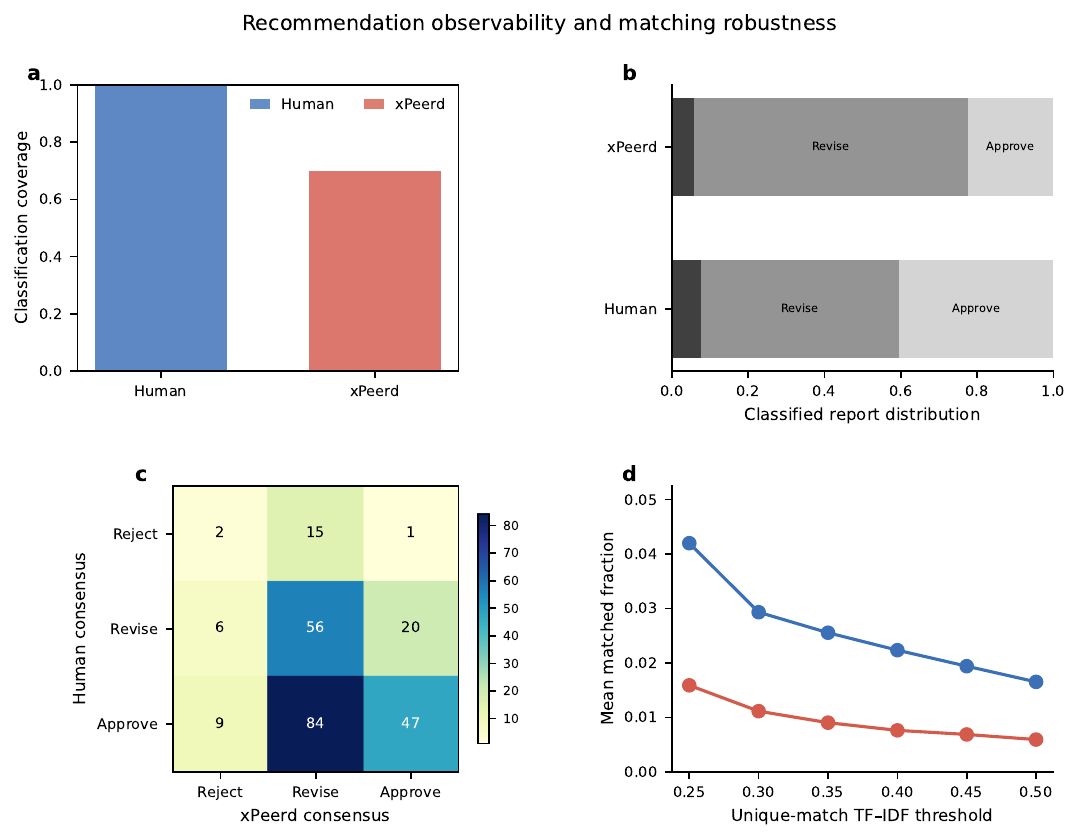}
\caption{Recommendation observability, normalized recommendation distributions, manuscript-level consensus, and lexical-matching sensitivity.}
\label{fig:s2-recommendations}
\end{figure}

\subsection{Quality criteria and reproducibility}
All 22 prespecified computational quality checks passed. The checks covered cohort counts, report balance, nonempty report text, concern-unit coverage, metric bounds, paired-row counts, sensitivity-threshold completeness, category reconciliation, recommendation auditing, and figure/output existence. The completion report also records association and agreement separately, documents lexical-grounding boundaries, and preserves the strict complete-case cohort size.

The benchmark is inspectable at record, report, concern-unit, manuscript-source, category, match, threshold, statistical-test, and figure levels. The quality criteria establish computational consistency, bounds, reconciliation, and artifact completeness. Construct validity and scientific correctness remain separate validation domains.

\section{Discussion}
The benchmark identifies a consistent but non-interchangeable pair of source profiles. \xpeer{} produced longer reports, more detector-recognized concerns, broader category representation, more explicit manuscript targets, and more requested actions. Human reports contained more explicit-rationale language, stronger lexical attestation to manuscript passages, slightly higher taxonomy-based scientific relevance, and lower mean within-source redundancy. Low cross-source lexical correspondence and recommendation agreement reinforce the conclusion that the two sources emphasize different observable aspects of a manuscript; they do not establish that either source is scientifically superior.

The profile differences reflect both review behavior and the measurement process. Longer reports provide more opportunities for concern extraction, while structured headings, action verbs, question forms, and task templates increase the visibility of \xpeer{} content to deterministic detectors and can promote systematic category coverage. Human reviewers may instead express causal reasoning implicitly, concentrate on fewer high-priority issues, and anchor judgments in domain knowledge without repeating manuscript language. Sixty-one reports yielded no extracted concern units, including 22 human and 39 \xpeer{} reports. Retaining these reports at zero preserves the paired design, but differential extraction failure may still affect source comparisons. Because no blinded expert-annotation subset was available to estimate source-specific precision, recall, category accuracy, or inter-annotator agreement, the concern-based results should be interpreted as measurements of declared textual observables rather than validated estimates of substantive critique.

Evidence from other tasks shows why these design factors should be separated. Prompt wording and reasoning-oriented fine-tuning change measured performance \cite{Wang20232609,Kim202312685}. Retrieval organization affects multihop reasoning \cite{Zhang202599376}, and specialized representations can matter in technical domains \cite{Hellert2025}. Clinical-decision research further illustrates the value of human expertise and multiparametric evaluation when assessing consequential model use \cite{Sblendorio2024}. These studies do not validate the present detectors, but they support reporting the prompt, retrieval, representation, and evaluation protocol explicitly.

The same distinction governs the alignment and agreement findings. TF--IDF and token overlap measure lexical attestation to manuscript chunks, not factual correctness, citation validity, or domain-grounded reasoning. Cross-source concern matching is threshold-sensitive and can conflate paraphrase, alternative decomposition, and genuinely different scrutiny. Expert adjudication is therefore required to separate useful additional concerns from redundant, irrelevant, or incorrect ones. Human reports are a manuscript-linked reference source, not a correctness gold standard, and neither report length nor agreement with one source is an adequate proxy for scientific validity. Recommendation findings require similar restraint because system labels were observable in 70.1\% of \xpeer{} reports and inference at that level is confined to the observable subset.

Cohort construction further limits the population to which the paired estimates apply. The strict cohort contains 271 of the 1,108 released records and 271 of the 802 records with exactly two human reports, corresponding to a 33.8\% complete-pair availability rate. This attrition may favor outputs with regular two-reviewer structure. Among the remaining 531 exact-two-human records, blank, recommendation-only, and partial-reviewer field states could reflect endpoint failure, incomplete generation, response-format variation, or parser allocation; the absence of raw transport responses prevents these mechanisms from being fully distinguished. The paired results consequently characterize complete cases rather than all attempted simulations. In addition, the F1000Research-linked manuscripts and reports represent an open post-publication setting, so transfer to anonymous pre-publication review, other venues, and different manuscript populations requires external validation.

Within those boundaries, the evidence supports using \xpeer{} as a pre-submission stress test. Its breadth can help researchers identify missing methodological detail, reporting omissions, unsupported interpretations, statistical issues, and presentation barriers before formal review. A disciplined workflow should verify the referenced manuscript location, examine the rationale, assess scientific relevance, prioritize severity, and record the revision decision for each high-impact concern. The lower explicit-reasoning score also identifies a product-development priority: consequential concerns should connect the observed issue, its methodological or evidential consequence, and the requested revision in a traceable sequence.

For editors, publishers, and conferences, the results justify evaluation of \xpeer{} as an additional scrutiny layer before or alongside human review, not as a transfer of editorial accountability. The system can provide a standardized methodological and reporting sweep and a structured issue inventory, but deployment should preserve inputs and outputs, disclose system use, distinguish generated recommendations from accountable editorial decisions, provide an appeal path, and monitor performance by field and manuscript type. Publishing proposals describe AI as support for human-led review and scientific discourse \cite{NatureBiomedEng2024665,Rowe202590}. Clinical decision-support research uses human expertise and multidimensional assessment to evaluate feasibility \cite{Sblendorio2024}. In education, one study evaluates automated subjective-answer scoring in terms of accuracy, fairness, and feedback \cite{Yuvasri2025648}; another reviews evidence about effects on human learning \cite{Li2025a}; and a third proposes a model for improving generated educational content \cite{Talaver2025149}. Retrieval-augmented safety work evaluates guidance for construction-risk management \cite{Baek2025}. These adjacent literatures support cautious evaluation within the tested context, not direct transfer of their performance findings to peer review.

The same qualification applies within medicine. Commentaries describe both promise and risk for language models in health care \cite{Yan20231657,Nelson2023408}. Task-specific evaluations examine dental diagnostic questions \cite{Danesh2025911}, prostate-cancer patient education \cite{Gibson2024}, and explanations of ophthalmology articles for patients \cite{Kianian2024}. Their heterogeneous populations and outcomes illustrate why usefulness claims must remain tied to a stated task and metric set. Here, the low lexical correspondence suggests differentiated focus, but editorial value depends on how many additional concerns survive factual, statistical, ethical, and domain-specific triage.

The review-withholding design supports a workflow-level claim: human-review text and recommendations were excluded from submitted inputs and joined only after \xpeer{} outputs had been persisted. It cannot exclude prior model exposure to publicly available manuscripts or review texts. Named competing systems were also outside the common-input experiment. Accordingly, the evidence supports claims about the evaluated system's documented operational access, public human-reference data, concern-level traceability, explicit attrition, and reproducible analysis, but not comparative superiority in scientific accuracy, cost, latency, throughput, or user experience.

The principal contribution is therefore an evidence infrastructure for stronger comparative testing: a public interface, a 1,108-record multidisciplinary resource, a same-manuscript paired cohort, 15,563 inspectable concern units, explicit attrition, uncertainty procedures, and machine-readable checks. Future systems can receive the same review-withheld manuscripts and be assessed with aligned cohort rules, concern extraction, profile metrics, category tests, overlap diagnostics, recommendation audits, cost and latency accounting, governance measures, and blinded expert adjudication. Agent orchestration with retrieval has been implemented in pharmacovigilance \cite{Choi2024}. Scientific knowledge-graph studies address semantic similarity and neuro-symbolic discovery \cite{Nguyen2025,Schmidt2024}, while retrieval research tests indexing strategies for multihop reasoning \cite{Zhang202599376}. Multi-agent GraphRAG offers another orchestration design in e-government \cite{Papageorgiou2025}. These systems do not provide peer-review evidence; they identify concrete retrieval and orchestration variants that a future common-input experiment could test. Contemporary peer-review studies contribute complementary multi-model and conference-scale evidence, but comparative ranking should rest on an identical-input, expert-adjudicated protocol.

\section{Conclusion}
This study presents a self-contained benchmark of \xpeer{} across operational behavior and same-manuscript human-reference evaluation. The operational component covers 352 valid simulation reports across disciplines and review modes. The public human-reference component contains 1,108 persisted records and a strict cohort of 271 manuscripts with two human and two usable \xpeer{} reports.

Under the declared extraction rules, \xpeer{} reports are longer, contain more concern units, name manuscript targets more often, represent more concern categories, and express more requested actions. Human reports contain more explicit rationale language, stronger lexical manuscript attestation, slightly higher taxonomy-based scientific relevance, and lower mean within-source redundancy. Lexical concern correspondence and recommendation agreement are low, establishing differentiated observable source profiles.

The public benchmark resource, explicit cohort accounting, inspectable concern-level measures, uncertainty analysis, and computational quality controls provide a transparent basis for independent replication and future common-protocol comparison. Scientific correctness and editorial utility remain questions for accountable expert judgment.

\section*{Data availability}
The study-level dataset, comprising 1,108 manuscript-level records, is available from Zenodo at \href{https://doi.org/10.5281/zenodo.21478076}{10.5281/zenodo.21478076} \cite{Knowdyn2026Benchmark}. The version 1.0.0 archive used for computational reproduction is preserved separately at \href{https://doi.org/10.5281/zenodo.21479700}{10.5281/zenodo.21479700} \cite{Knowdyn2026BenchmarkVersion}. The former DOI identifies the evolving study record, whereas the latter provides a version-specific reference to the exact reproducibility archive. Reports used in the operational component were provided by KNOWDYN under permission. Aggregate findings derived from these reports are presented in this article, but the source reports remain under the provider's control and are not publicly released.

\section*{Code availability}
The benchmark-construction and TRACE-R notebooks, supporting reproducibility materials, and machine-readable outputs are publicly available at \href{https://github.com/khalid-saqr/xPeerd_Evaluation}{github.com/khalid-saqr/xPeerd\_Evaluation}. Repository commit \texttt{99e602873ddb1f7dca8a08d8aa05979e1fce643e} is the frozen code reference for this article \cite{xPeerEvaluationRepo2026}. Authentication credentials are excluded from the notebooks, archives, logs, and repository history. The article package includes the complete LaTeX source and bibliography, all empirical figure files, and the TikZ source used to generate the benchmark diagrams.

\section*{Ethics statement}
The study analyzed existing scholarly manuscripts and peer-review records and did not recruit or intervene with human participants. The human-reference materials were obtained from the openly licensed Re3-Sci2.0 resource.

\section*{Competing interests}
The author holds a controlling share of KNOWDYN LTD, the owner of \xpeerd{} and its commercial operations. Public data, released code, explicit exclusions, hashes, claim controls, independent replication, and external adjudication form the conflict-management framework.

\section*{Acknowledgements}
The author acknowledges the creators of Re3-Sci and Re3-Sci2.0 and the authors and reviewers whose openly licensed manuscript and review records made the human-reference benchmark possible.

\printbibliography
\end{document}